\documentclass{article}

\usepackage{arxiv}

\usepackage[utf8]{inputenc} 
\usepackage[T1]{fontenc}    
\usepackage{hyperref}       
\usepackage{url}            
\usepackage{booktabs}       
\usepackage{amsfonts}       
\usepackage{nicefrac}       
\usepackage{microtype}      
\usepackage{lipsum}
\usepackage{graphicx}

\usepackage{caption} 
\usepackage{algorithm}
\usepackage{algpseudocode}
\usepackage{amsmath}
\usepackage{amssymb}
\usepackage{enumitem}
\usepackage{array}
\usepackage{multirow}
\usepackage{tikz}

\usepackage{newfloat}
\usepackage{listings}
\newfloat{listing}{tb}{lst}{}
\floatname{listing}{Listing}

\graphicspath{ {./images/} }

\title{PhysX-CoT: Structured Physical Reasoning from a Single Image to Simulation-Ready 3D Assets}

\author{
 Jie Huang, Xiaohe Li$^{*}$, Jiahao Li, Fangli Mou\\
 	the Key Laboratory of Target\\
 	Cognition and Application Technology\\
	Aerospace Information Research Institute\\
	Chinese Academy of Sciences \\
  \texttt{lixiaohe@aircas.ac.cn} \\
   \And
 Chen Qian \\
  School of Artificial Intelligence\\
  Shanghai Jiaotong University \\
  \texttt{qianc@sjtu.edu.cn} \\
  \And
  Yuqiang Fang \\
  Space Engineering University\\
  \texttt{fangyuqiang@nudt.edu.cn} \\
  \And
 Junhao Fan, Kaixin Zhang, Zide Fan$^{*}$ \\
the Key Laboratory of Target\\
Cognition and Application Technology\\
Aerospace Information Research Institute\\
Chinese Academy of Sciences\\
\texttt{fanzd@aircas.ac.cn}}

\begin{document}
\maketitle
\begin{abstract}
	Simulation-ready 3D assets are central to robotics and embodied AI. Generating them from a single image is usually framed as a vision-language model that emits a serialized asset for a decoder to turn into geometry and physical fields, leaving the image-to-3D reasoning implicit. We argue the limiting factor is this output-centric view: part placement and local shape are entangled in one global-coordinate token stream, and the intermediate physical states are never exposed for supervision, conditioning, or verification. PhysX-CoT instead casts single-image asset generation as an explicit structured physical reasoning process, an ordered and machine-parseable trajectory of part-level states covering decomposition, 2D and 3D grounding, relations, coarse geometry, and surface cues that we separately supervise, use to condition geometry, and treat as reward targets. Geometry is factorized so that 3D boxes carry placement and local codes carry shape, and CoT-aligned GRPO optimizes parse validity, grounding, geometry, placement, and physical consistency. Under a unified protocol that retrains all learned baselines on the same backbone, data, and frozen decoder, PhysX-CoT outperforms the closest full-task baseline across geometry, scale, and physical-attribute metrics. Oracle, token-matched, and state-order controls show the explicit states are functional rather than cosmetic, and in Unreal Engine~5 the generated assets parse, collide, and articulate at high validity.
\end{abstract}

\section{Introduction}

\begin{figure}[t]
	\centering
	\includegraphics[width=\linewidth]{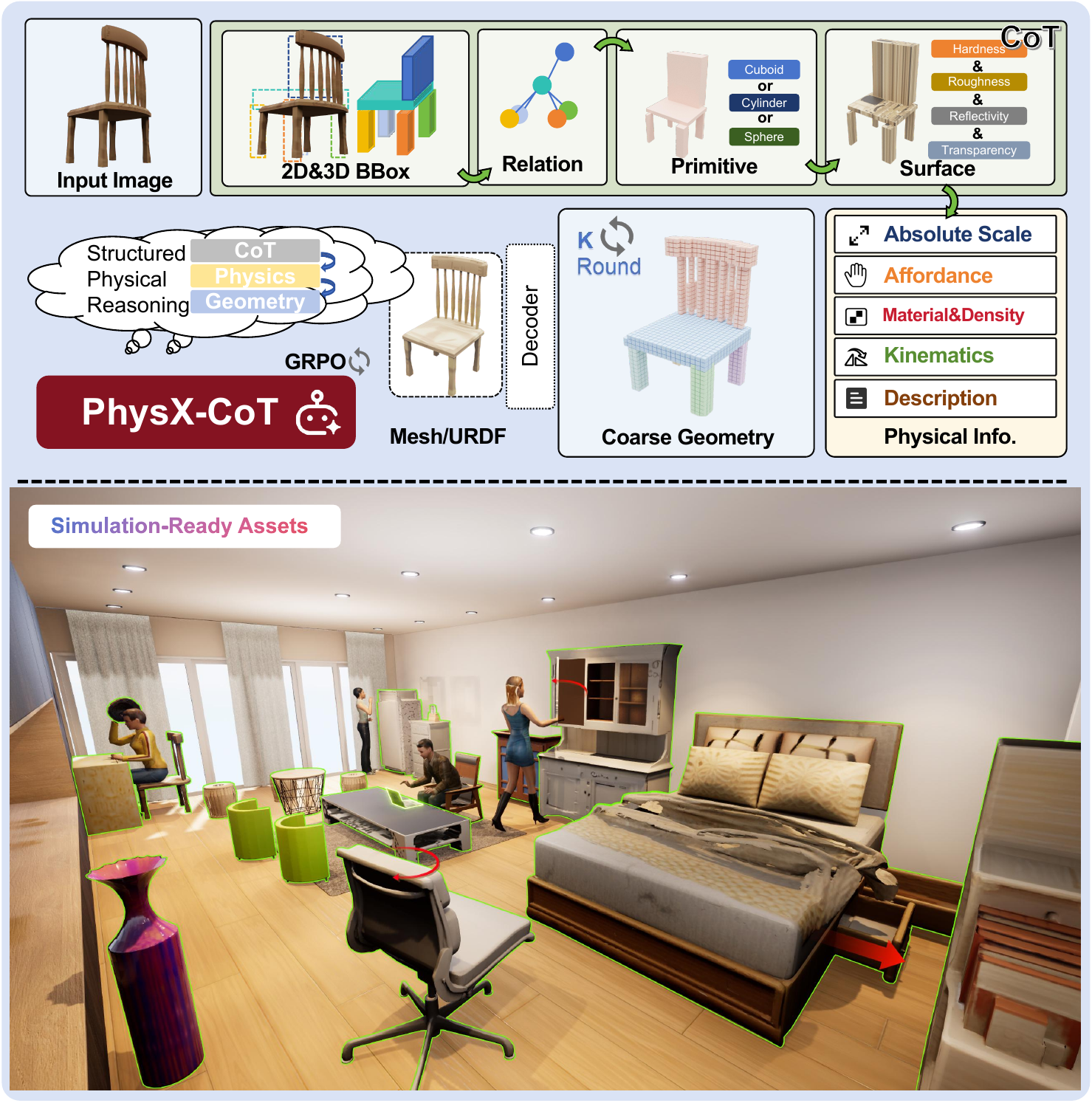}
	\caption{PhysX-CoT recasts single-image, simulation-ready 3D generation as a \emph{structured physical reasoning} process: an ordered, inspectable chain of CoT states (2D/3D boxes, relations, primitives, surface cues) and object-level physical attributes (scale, material, affordance, kinematics), decoded into mesh/URDF assets that populate a simulation-ready scene (bottom).}
	\label{fig:teaser}
\end{figure}

Converting images into simulation-ready 3D assets could provide a scalable bridge from abundant visual data to interactive virtual worlds, with broad implications for robotics, embodied AI, and physical simulation. Neural and explicit scene representations, including neural radiance fields and 3D Gaussian splatting, have enabled high-fidelity reconstruction and rendering~\cite{mildenhall2020nerf,kerbl2023gaussian}, while generative models have learned to synthesize textured shapes and distill 2D diffusion priors into 3D~\cite{gao2022get3d,poole2023dreamfusion}. Together, these advances have made it possible to generate plausible 3D geometry from a single image within seconds: view-conditioned and multi-view diffusion models provide novel-view priors~\cite{liu2023zero123,long2024wonder3d}, large feed-forward models reconstruct 3D assets in a single pass~\cite{hong2024lrm}, and structured-latent generators scale image-conditioned synthesis~\cite{xiang2024trellis}. However, visually plausible 3D content is not necessarily physically usable. Existing methods largely optimize how an object looks, leaving unresolved the metric scale, part hierarchy, physical materials, collision geometry, and articulation required for interaction in a physics engine.

This gap matters for robotics and embodied AI, where assets must behave in a simulator, not merely look correct. Interactive environments rely on part-based, articulated assets with kinematic annotations \cite{xiang2020sapien}, and even visually faithful reconstructions can topple or fall apart under gravity and contact once simulated \cite{guo2024physicalcompatible}. In manipulation and policy learning, a wrong joint axis or an unstable contact can quietly invalidate an entire training rollout. Recent simulation-ready generators target this bar directly: PhysX-3D contributes physics-grounded annotations and a feed-forward generator \cite{cao2025physx3d}, while PhysX-Anything uses a vision--language model (VLM) to produce geometry, articulation, and physical attributes from a single image \cite{cao2025physxanything}. These systems mark real progress, but they predict the physical outputs while leaving the reasoning that links image evidence to each part implicit.

Two problems follow from this output-centric view. First, geometry is written in one shared, object-level coordinate frame, whether as global voxels or as serialized tokens, so the same tokens must specify both where a part sits and what shape it takes. We call this where--shape coupling: once a predicted position drifts, the shape decodes into the wrong region, and for thin or topologically complex parts the asset fractures, drifts, or collapses in simulation. Second, the grounding and relational states a human would reason through, such as which parts exist and how they connect and move, are neither represented, supervised, nor verified on their own. Language-model geometry serializers make this concrete, since emitting a mesh as one token stream couples structure with position and exposes no checkable intermediate state \cite{wang2024llamamesh,chen2024meshanything,ye2025shapellmomni}.

Simply prompting the VLM to name these fields changes little in our experiments: a chain of states helps only when it is supervised and verifiable, not a cosmetic prefix. Process supervision can improve reasoning by rewarding verifiable intermediate steps rather than only final outcomes \cite{lightman2023verify}, while grounded-reasoning work turns boxes, regions, and layouts into explicit structure \cite{cheng2024spatialrgpt,feng2023layoutgpt}. PhysX-CoT brings this view to asset generation. It exposes the VLM stage as an ordered, machine-parseable trajectory of part-level physical states that we separately supervise, use to condition local geometry, and optimize with process-level rewards through a parser and verifier, and it factorizes geometry so that a 3D box carries placement and a local-frame code carries shape. To isolate this interface, we reuse the downstream decoder and evaluation protocol of PhysX-Anything unchanged \cite{cao2025physxanything} and change only the VLM stage, probing the intermediate states directly with oracle and perturbation controls. Concretely (Figure~\ref{fig:teaser}), PhysX-CoT comprises a Structured Physical Chain-of-Thought over part states, Position-Factorized Local Geometry (a 3D box for placement, a local-frame code for shape), and CoT-aligned GRPO. Here CoT names this ordered, parseable state trajectory, not a natural-language rationale, and unlike staged bbox-then-shape decoding, the order itself is supervised and shown functional.

The main contributions of this paper are as follows:
\begin{itemize}
	\item We introduce PhysX-CoT, which recasts single-image simulation-ready 3D generation as an explicit structured physical reasoning process: it exposes the VLM stage as an ordered, machine-parseable trajectory of part-level physical states that are separately supervised, condition geometry, and serve as reward targets.
	\item We propose position-factorized local geometry, which separates a part's placement from its local shape so that each is represented, supervised, and rewarded on its own, and CoT-aligned GRPO, which optimizes parse validity, grounding, local geometry, placement, and physical consistency through a parser and verifier.
	\item We further design an evaluation that probes the reasoning process, not only the final assets: under the unified PhysX-CoTA protocol, oracle, perturbation, and state-order controls and a standardized Unreal Engine~5 benchmark show that the ordered states are functionally used rather than merely cosmetic.
\end{itemize}

\section{Related Work}

\subsection{3D Generative Models}

Research on 3D generation has advanced along a clear trajectory. Early models learned shape distributions with voxel, implicit, and GAN representations \cite{wu2016learning,mescheder2019occupancy,gao2022get3d}, and optimization-based text-to-3D later distilled 2D diffusion priors to reach far higher fidelity, at the cost of slow per-shape optimization \cite{poole2023dreamfusion,lin2023magic3d,wang2023prolificdreamer}. To remove that cost, feed-forward and multi-view methods predict geometry directly: view-conditioned and multi-view diffusion enforce cross-view consistency \cite{liu2023zero123,long2024wonder3d,shi2024mvdream}, feed-forward reconstruction models map images directly to 3D representations \cite{liu2023one2345,hong2024lrm,tang2024lgm,xu2024instantmesh}, and structured-latent generators support scalable image-conditioned 3D synthesis \cite{xiang2024trellis}. A parallel line lets autoregressive language models emit meshes as token sequences \cite{siddiqui2024meshgpt,chen2024meshxl,chen2024meshanything}. Throughout this progression the objective has stayed visual or geometric plausibility, not the physical structure a simulator consumes, so scale, articulation, and material properties are absent or deferred to post-processing.

\begin{figure*}[!t]
	\centering
	\includegraphics[width=\textwidth]{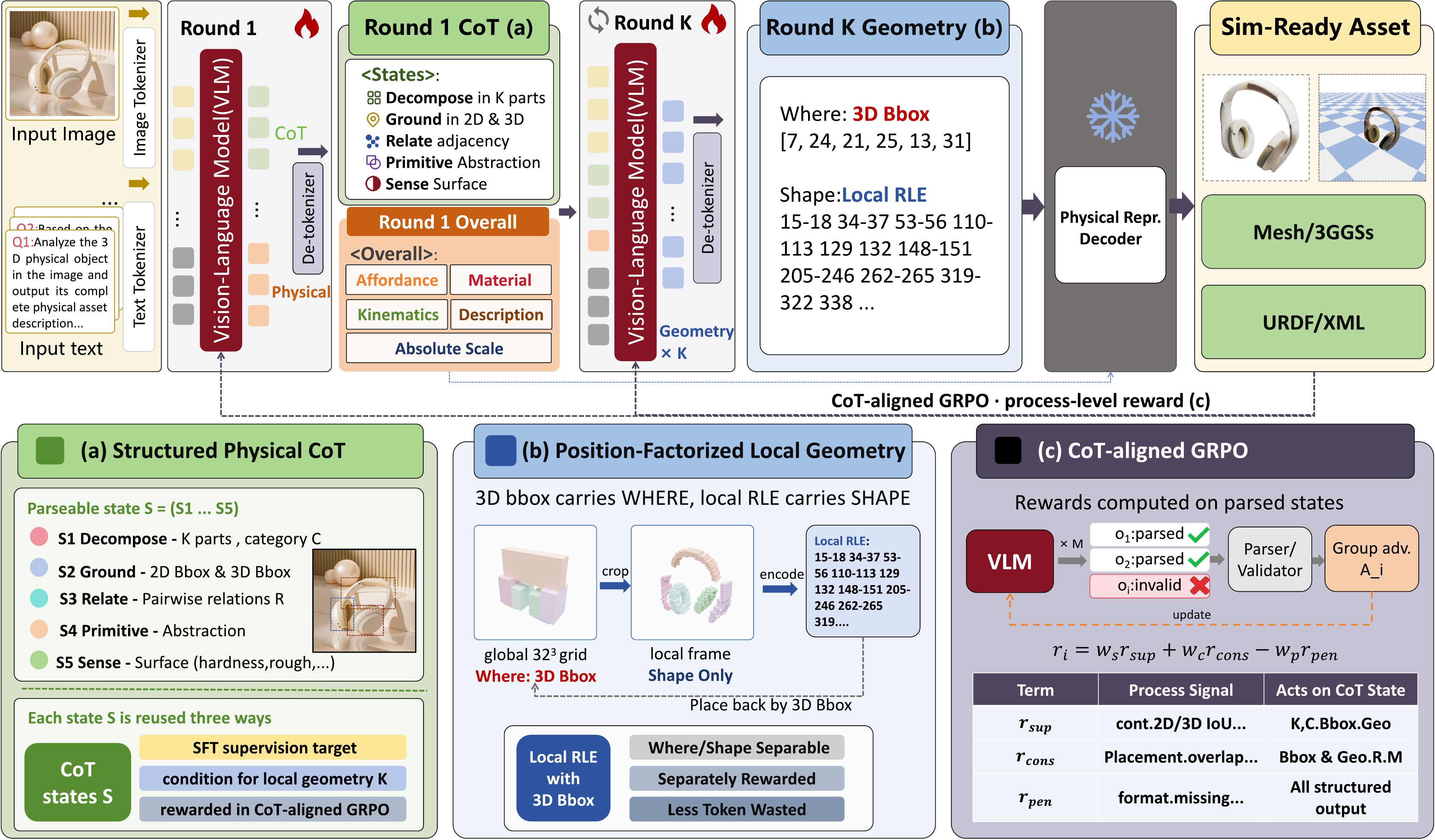}
	\caption{Overview of PhysX-CoT. Round~1 emits the structured physical CoT and object-level attributes; Rounds~$2\dots K$ emit per-part geometry; a frozen decoder yields the simulation-ready asset (mesh/URDF); CoT-aligned GRPO applies process-level rewards. \textbf{(a) Structured Physical CoT}: an \emph{ordered}, parseable state trajectory $S=(K,C,\{s_k\})$ (decompose, 2D/3D ground, relate, primitive, sense surface); each state is an SFT target, a geometry condition, and a reward target. \textbf{(b) Position-Factorized Local Geometry}: a 3D box carries \emph{where} and a local-frame RLE carries \emph{shape}, cropped, encoded, and placed back by the box, with separate local-shape and global-placement rewards. \textbf{(c) CoT-aligned GRPO}: parsed/verified candidates yield process-level rewards driving group-relative updates.}
	\label{fig:overview}
\end{figure*}

\subsection{Articulated and Physical 3D Asset Generation}

Closing this gap has meant progressively raising what a generated asset must contain. Articulated modeling first added part- and interaction-aware structure: simulators and datasets such as SAPIEN supply part and kinematic annotations \cite{xiang2020sapien}, and generators build movable objects or URDF descriptions from images, generated code, point clouds, or existing meshes \cite{chen2024urdformer,liu2024singapo,le2024articulateanything,li2025urdfanything,li2025particulate}. Many of these introduce structured intermediates, such as part-connectivity graphs or generated code, yet use them inside a pipeline rather than as separately supervised, image-grounded states. Physical asset generation then extended the target beyond geometry to physical attributes: PhysX-3D grounds assets in scale, material, affordance, and kinematics \cite{cao2025physx3d}, and PhysX-Anything produces simulation-ready geometry, articulation, and attributes from a single image with a VLM \cite{cao2025physxanything}, while related work enforces static stability or plans an object-level physical blueprint before geometry \cite{guo2024physicalcompatible,yang2026physforge}. Complementary work addresses physically plausible composition and stable scene reconstruction \cite{yan2024phycage,ma2026rest3d}, scalable simulation-ready assets and worlds \cite{feng2025seed3d,wang2025embodiedgen}, and part-level or animatable modeling \cite{chen2024partgen,huang2026anigen}; SimuScene instead benchmarks code generation for physical-scenario simulation \cite{wang2026simuscene}. PhysX-Omni further unifies rigid, deformable, and articulated generation using an explicit template-based RLE representation for higher-resolution geometry and the broader PhysXVerse dataset \cite{cao2026physxomni}. In contrast, our focus is not broader asset coverage but an ordered image-to-asset state trajectory whose intermediate variables are separately supervised, evaluated, and optimized with process rewards.

\subsection{VLM-based 3D Generation}

Our approach sits at the intersection of this trajectory and a second one: the use of vision-language models for 3D. Language models first learned to emit geometry directly, serializing meshes or 3D tokens for generation and editing \cite{wang2024llamamesh,chen2024meshanything,fang2025meshllm,ye2025shapellmomni}, and PhysX-Anything carried this to simulation-ready assets through multi-round VLM prediction \cite{cao2025physxanything}, with concurrent MLLM methods decomposing meshes into sim-ready articulated assets \cite{zhang2026simart}. In parallel, 3D and point-cloud features were injected into LLMs for captioning, grounding, and question answering \cite{hong2023threedllm,xu2024pointllm,qi2024gpt4point}, and VLMs were given spatial, grounded reasoning over boxes, regions, layouts, and programs \cite{chen2024spatialvlm,cheng2024spatialrgpt,feng2023layoutgpt,suris2023vipergpt}. Yet most of these systems either emit geometry directly or keep their reasoning as free-form language, so the intermediate states are rarely a fixed schema and almost never separately supervised, executed, or rewarded. Even where explicit rationales appear, as in multimodal chain-of-thought or 3D reasoning fine-tuning, they stay natural-language explanations rather than machine-checkable geometry \cite{zhang2023multimodalcot,chen2026pointllmr}. This progression connects explicit reasoning traces \cite{wei2022cot} with verifiable process supervision \cite{lightman2023verify} and reward-based optimization through GRPO \cite{shao2024deepseekmath}. PhysX-CoT departs from it by making the VLM's image-to-3D reasoning a supervised and verifiable object, so the reasoning itself, not only the final asset, is trained and checked.

\section{Method}

\subsection{Overview}

Given a single RGB image $I$ and prompt $q$, we generate a simulation-ready asset $A$. PhysX-CoT (Figure~\ref{fig:overview}) modifies the VLM-stage representation, geometry tokenization, and training objective, reusing the downstream decoder unchanged. We write the VLM-stage output as
\begin{equation}
	Y=\left(S,P,\{G_k\}_{k=0}^{K-1}\right), \quad
	A=\mathcal{D}_{\mathrm{phys}}(Y,I),
\end{equation}
where $S$ is the structured physical CoT, $P$ the object-level physical description, $G_k$ the local coarse-geometry sequence for part $k$, and $\mathcal{D}_{\mathrm{phys}}$ the physical decoder.

Generation uses a $1{+}K$-round decomposition: the first round predicts $S$ and $P$, then part-specific prompts $q_k$ built from $S$ generate each $G_k$ in turn:
\begin{equation}
	p_{\theta}(Y|I,q)=p_{\theta}(S,P|I,q_0)
	\prod_{k=0}^{K-1}p_{\theta}(G_k|I,q_k,S,P,s_k).
\end{equation}

\subsection{Structured Physical CoT Schema}

Structured Physical CoT organizes these intermediate variables into machine-parseable states:
\begin{equation}
	S=(K,C,\{s_k\}_{k=0}^{K-1}),\quad
	s_k=(B_k^{2D},B_k^{3D},R_k,H_k,M_k).
\end{equation}
Here $K$ and $C$ are the part count and semantics, $B_k^{2D}$ and $B_k^{3D}$ the 2D and axis-aligned 3D boxes, $R_k$ the inter-part relations, $H_k$ a coarse-geometry prior, and $M_k$ surface cues; object-level $P$ holds global information (scale, category, density, joint grouping, function). $S$ is the part-level interface that conditions $G_k$ and receives rewards; full field definitions are in the appendix.

In training, $S$ comes from part annotations, projection/voxel boxes, and joint relations, and $G$ from RLE-encoding cropped part voxels; open-ended fields are VLM-proposed and human-verified (appendix).
\subsection{Position-Factorized Local Geometry}

PhysX-CoT uses $B_k^{3D}$ for where and $G_k$ for shape. We say \emph{factorized}, not fully decoupled: local geometry is still conditioned on the predicted box, but placement and shape become separately supervised and rewarded. For the $k$-th part,
\begin{equation}
	B_k^{3D}=(x_{\min}^k,x_{\max}^k,y_{\min}^k,y_{\max}^k,z_{\min}^k,z_{\max}^k)
\end{equation}
is defined in a $32^3$ object voxel grid. Each occupied voxel inside the box is mapped to a local linear index, and the sorted indices are RLE-compressed into $G_k$ (wrapped by \texttt{<geometry\_l\_k>}); the local index space adapts to each part's dimensions rather than resampling to a common cube. The exact index map and parser validity rules are in the appendix.

Local RLE describes only the shape inside the box, so geometry tokens no longer carry both localization and shape and are markedly more compact: under the same tokenizer they use about $58\%$ fewer geometry tokens than the global encoding, freeing capacity for the structured states. At reconstruction, $G_k$ is decoded and placed back via $B_k^{3D}$; overlapping boxes are resolved at the coarse-voxel stage by assigning each shared voxel to the higher-confidence part (physics-based resolution is future work).

\subsection{CoT-Aligned Training}

PhysX-CoT trains in two stages, SFT then GRPO \cite{shao2024deepseekmath}. SFT learns the mapping from the image to $Y=(S,P,\{G_k\})$:
\begin{equation}
	\mathcal{L}_{\mathrm{SFT}}=-\sum_{t=1}^{T}\log p_{\theta}(y_t|y_{<t},I,q).
\end{equation}
The loss covers CoT states, object-level descriptions, and local RLE tokens; targets $B^{3D}$ and $G$ come from cropping and encoding part voxels, so supervision and rewards share one coordinate definition.

PhysX-CoT is built on Qwen3-VL-8B-Instruct and trained on $4{\times}$A800 GPUs. The GRPO stage starts from the SFT checkpoint and tunes LoRA on the LM projections only (appendix). For each prompt we sample $N{=}4$ candidates, score each with a scalar sequence reward $r_i$, and use the group-relative advantage $A_i = r_i - \tfrac{1}{N}\sum_{j=1}^{N} r_j$. The policy is updated with a clipped policy-ratio objective and a KL anchor to the frozen SFT reference $\pi_{\mathrm{ref}}$:
\begin{equation}
	\begin{aligned}
		\mathcal{L}_{\mathrm{GRPO}}=-\mathbb{E}_{i,t}\big[&\min(\rho_{i,t}A_i,\,\mathrm{clip}(\rho_{i,t},1{-}\epsilon,1{+}\epsilon)A_i)\big]\\
		&+\,\beta\,D_{\mathrm{KL}}(\pi_\theta\,\|\,\pi_{\mathrm{ref}}),
	\end{aligned}
\end{equation}
where $\rho_{i,t}$ is the token-level policy ratio and the sequence reward $r_i$ is assigned to all tokens of candidate $i$, with $\beta{=}0.02$ and $\epsilon{=}0.2$. The scalar reward aggregates four process terms and a penalty,
\begin{equation}
	r_i=\mathrm{clip}_{[-2,2]}\!\Big(\tfrac{\sum_{m\in\mathcal{M}}\lambda_m R_m}{\sum_{m\in\mathcal{M}}\lambda_m}-\lambda_{\mathrm{pen}}P\Big),
\end{equation}
where $\mathcal{M}$ collects the applicable reward terms scoring localization, coarse primitives, local/global voxel quality (the RL core), and physical/joint validity, and $P$ penalizes format, range, and dependency violations. All rewards use only training-split annotations, and evaluation uses unseen objects and views (formulas in the appendix).

\subsection{Inference and Asset Reconstruction}

At inference the model generates $S,P$; the parser extracts the states and boxes (rejecting or resampling invalid outputs), generates each $G_k$, maps it back via $B_k^{3D}$, and feeds the result with $P$ to the decoder for meshes, attributes, joints, and simulation-compatible structures.

\section{Experiments}


We report final-asset comparisons and module ablations under the standard protocol, process-level diagnostics (state accuracy, controlled interventions, and a CoT-state linkage), and in-the-wild and UE5 generalization.

\begin{table*}[!t]
	\centering
	\small
	\setlength{\tabcolsep}{4pt}
	\begin{tabular}{l|ccc|ccccc}
		\toprule
		\multirow{2}{*}{Methods} & \multicolumn{3}{c|}{Geometry} & \multicolumn{5}{c}{Physical Attributes} \\
		& PSNR$\uparrow$ & CD$\downarrow$ & F-score$\uparrow$ & Absolute scale$\downarrow$ & Material$\uparrow$ & Affordance$\uparrow$ & Kinematic (VLM)$\uparrow$ & Description$\uparrow$ \\
		\midrule
		URDFormer & 12.56 & 0.334 & 0.192 & -- & -- & -- & 0.31 & -- \\
		Articulate-Anything & 14.82 & 0.321 & 0.024 & -- & -- & -- & 0.38 & -- \\
		PhysXGen & 16.42 & 0.282 & 0.056 & 28.02 & 8.49 & 8.53 & 0.46 & 10.17 \\
		PhysX-Anything & 19.10 & 0.095 & 0.256 & 12.43 & 14.00 & 15.17 & 0.53 & 18.12 \\
		\midrule
		\textbf{PhysX-CoT (Ours)} & \textbf{21.33} & \textbf{0.041} & \textbf{0.480} & \textbf{6.57} & \textbf{17.09} & \textbf{17.69} & \textbf{0.73} & \textbf{21.80} \\
		\bottomrule
	\end{tabular}
	\caption{Main results on the PhysX-CoTA test set. ``--'' indicates that a method does not directly output the corresponding capability and is not included in ranking or penalization for that metric.}
	\label{tab:main_results}
\end{table*}
\begin{figure*}[!t]
	\centering
	\includegraphics[width=\textwidth]{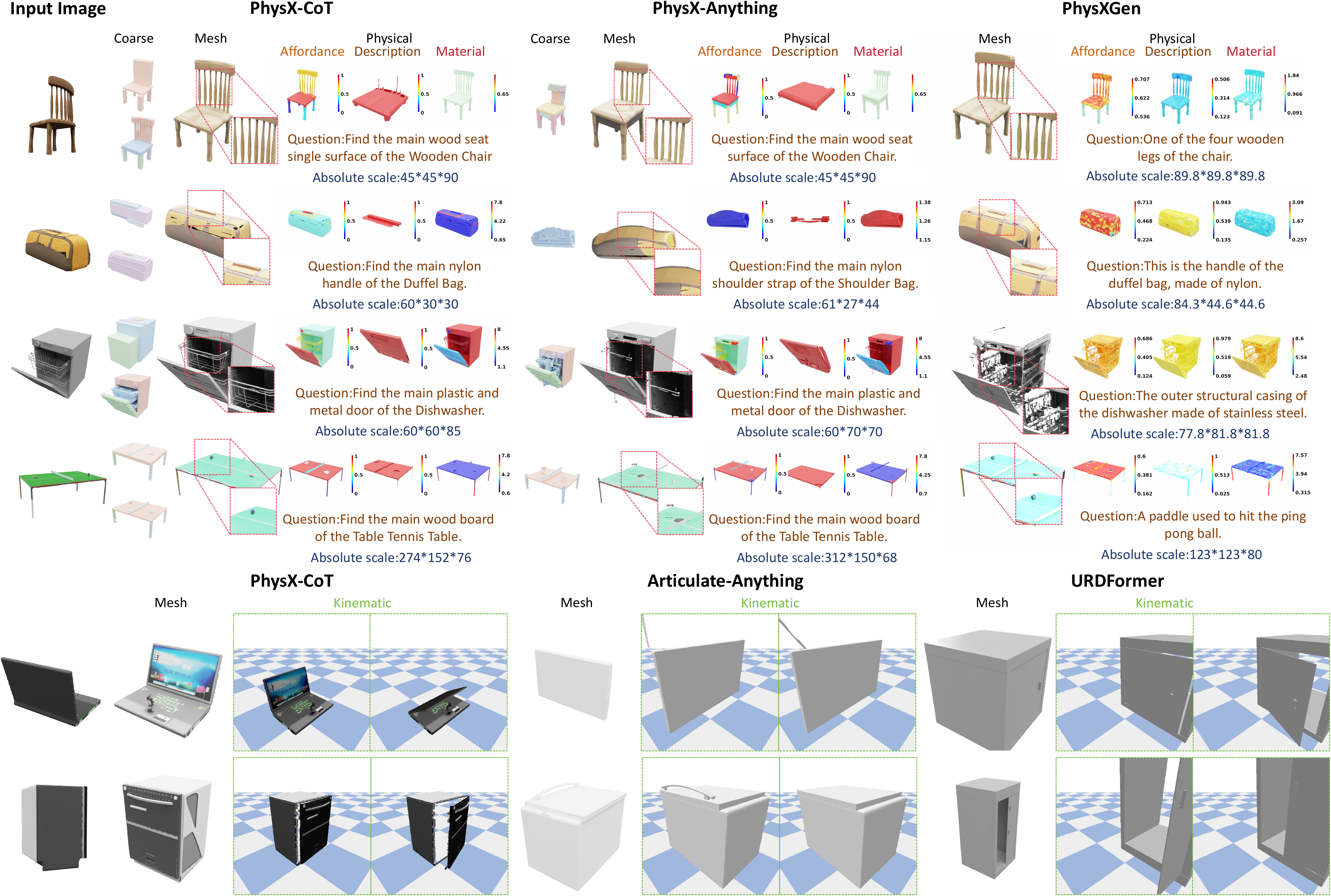}
	\caption{Main qualitative comparison on PhysX-CoTA across geometry, physical-property, and kinematic fields. Against PhysX-Anything, PhysXGen, Articulate-Anything, and URDFormer, PhysX-CoT recovers thin/small parts, absolute scale, and joint structure more reliably.}
	\label{fig:main_results}
\end{figure*}
\begin{figure*}[!t]
	\centering
	\includegraphics[width=\textwidth]{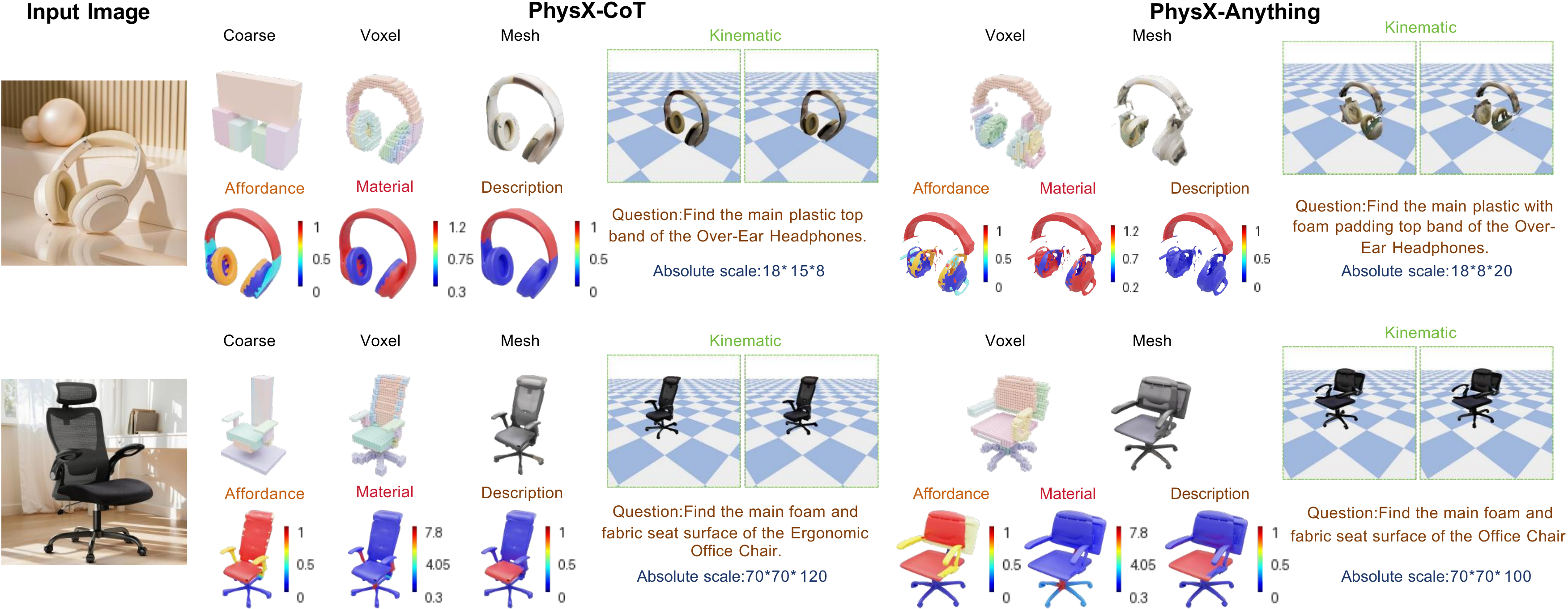}
	\caption{In-the-wild generalization (qualitative): on real photographs, PhysX-CoT recovers parts, thin/small structures, scale, and joints under varied backgrounds and viewpoints.}
	\label{fig:inthewild}
\end{figure*}
\begin{figure*}[!t]
	\centering
	\includegraphics[width=\textwidth]{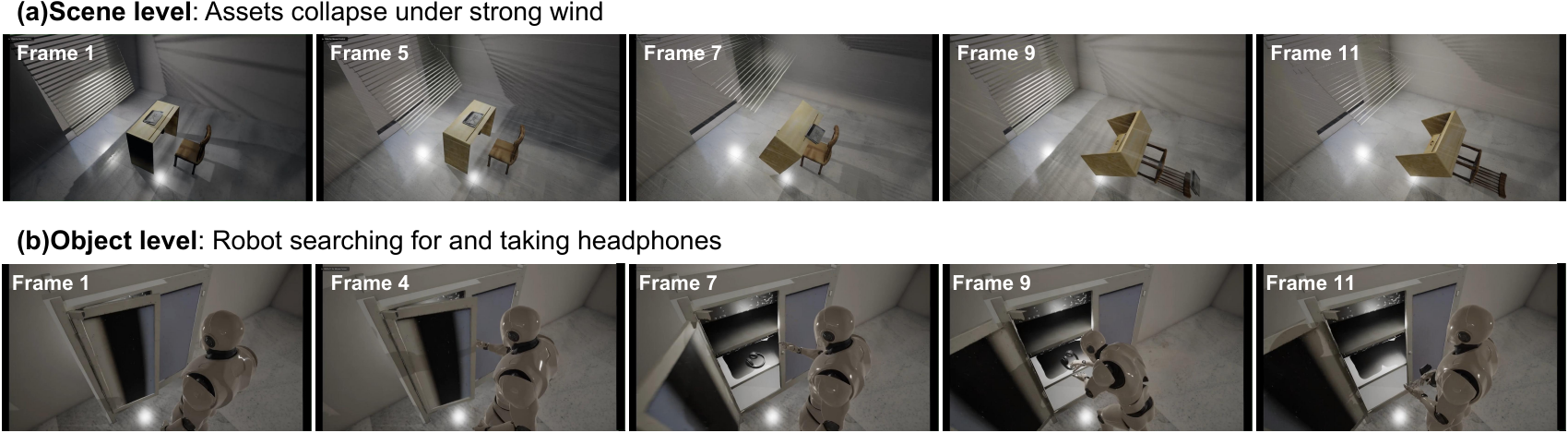}
	\caption{Simulation execution in Unreal Engine~5. \textbf{(a) Scene level}: under a strong wind field, multiple generated assets interact and topple as physical rigid bodies across frames. \textbf{(b) Object level}: a robot searches for and picks up a generated pair of over-ear headphones, showing the asset loads and is graspable.}
	\label{fig:simulation_results}
\end{figure*}

\subsection{Experimental Setup}

\textbf{Dataset.} PhysX-CoTA augments PhysXNet with Structured Physical CoT content (no new assets) \cite{cao2025physx3d}; splits are by object ID, giving $669{,}141$ single-image samples over $32{,}047$ objects. Open-ended fields are VLM-proposed, human-verified (appendix).

\textbf{Baselines.}
URDFormer and Articulate-Anything (articulated), PhysXGen (PhysX-3D's generator), and PhysX-Anything (closest full-task); ablations add PhysX-Anything+CoT Prompt. Unsupported fields are ``--'' (excluded from ranking). For fair attribution, all learned baselines are retrained under matched conditions (same backbone where applicable, same data and object-ID splits, same frozen decoder, matched budget), so only the VLM-stage representation and objective differ; reported numbers follow this normalized protocol and are not directly comparable across datasets.

\textbf{Metrics.}
Geometry uses PSNR, CD, and F-score, with CD and F-score primary: CD on $10{,}000$ samples, object-scale-normalized, without ICP; surface F-score at $\tau{=}0.01$ on the normalized post-decoder mesh; PSNR over multi-view renders. Scale is the relative dimension error (\%). Adapted from PhysX-3D \cite{cao2025physx3d}, Material is an MAE-based similarity between predicted and reference attributes, Affordance is an affordance-ranking score, and Description is a prompt cosine-similarity score; we report all three as higher-is-better similarities, with exact formulas and normalization in the appendix. Only Kinematic (joint type, axis, range) is VLM-judged, matching expert labels at Cohen's $\kappa{=}0.79$ on a human-annotated subset, and we additionally report the objective joint-axis error in Table~\ref{tab:state_main}; full definitions are in the appendix.

\begin{table*}[!t]
	\centering
	\small
	\setlength{\tabcolsep}{3pt}
	\begin{tabular}{l|ccc|ccccc}
		\toprule
		\multirow{2}{*}{Method Variant} & \multicolumn{3}{c|}{Geometry} & \multicolumn{5}{c}{Physical Attributes} \\
		& PSNR$\uparrow$ & CD$\downarrow$ & F-score$\uparrow$ & Absolute scale$\downarrow$ & Material$\uparrow$ & Affordance$\uparrow$ & Kinematic (VLM)$\uparrow$ & Description$\uparrow$ \\
		\midrule
		PhysX-Anything & 19.10 & 0.095 & 0.256 & 12.43 & 14.00 & 15.17 & 0.53 & 18.12 \\
		PhysX-Anything+CoT Prompt & 19.47 & 0.088 & 0.261 & 10.77 & 14.03 & 15.96 & 0.53 & 18.22 \\
		PhysX-Anything+Local Geo. & 20.11& 0.059& 0.352& 10.68& 14.08& 16.66& 0.59& 18.22\\
		\midrule
		PhysX-CoT-SFT & 19.62 & 0.064 & 0.330 & 6.85 & 16.92 & 17.11& 0.58& 21.03\\
		+ Local Geo. & 20.85 & 0.051 & 0.434 & 6.81 & 17.02 & 17.32& 0.63& 21.47\\
		\textbf{+ GRPO (Full)} & \textbf{21.33} & \textbf{0.041} & \textbf{0.480} & \textbf{6.57} & \textbf{17.09} & \textbf{17.69} & \textbf{0.73} & \textbf{21.80} \\
		\bottomrule
	\end{tabular}
	\caption{Ablation study. All variants use the same simulation-ready asset evaluation metrics as the main experiments.}
	\label{tab:ablation}
\end{table*}

\begin{table}[!t]
	\centering\small
	\setlength{\tabcolsep}{3pt}
	\begin{tabular}{lccccc}
		\toprule
		Reward & CD$\downarrow$ & F-sc.$\uparrow$ & Scale$\downarrow$ & Kin.$\uparrow$ & Parse$\uparrow$ \\
		\midrule
		Full GRPO & \textbf{0.041} & \textbf{0.480} & \textbf{6.57} & \textbf{0.73} & \textbf{96.4} \\
		\quad w/o $R_{\mathrm{loc}}$ & 0.047 & 0.459 & 7.34 & 0.70 & 95.8 \\
		\quad w/o $R_{\mathrm{detail}}$ & 0.054 & 0.441 & 6.78 & 0.68 & 95.6 \\
		\quad w/o $R_{\mathrm{phys}}$ & 0.044 & 0.472 & 7.09 & \underline{0.62} & 95.7 \\
		\quad w/o format penalty & 0.049 & 0.451 & 6.91 & 0.67 & \underline{87.1} \\
		\bottomrule
	\end{tabular}
	\caption{Reward ablation of CoT-aligned GRPO (\emph{Parse}: parse-valid rate, \%); removing a term most hurts the metric it targets (worst per column underlined).}
	\label{tab:reward_ablation}
\end{table}

\begin{table}[!t]
	\centering\small
	\setlength{\tabcolsep}{3pt}
	\begin{tabular}{lccccc}
		\toprule
		Method & 3D IoU$\uparrow$ & Rel.\ F1$\uparrow$ & Prim.$\uparrow$ & Joint$\uparrow$ & Axis$^\circ\!\downarrow$ \\
		\midrule
		PhysX-Anything & 0.37 & 0.39 & 0.48 & 0.53 & 35.6 \\
		\;+CoT Prompt & 0.39 & 0.41 & 0.50 & 0.54 & 34.8 \\
		PhysX-CoT-SFT & 0.56 & 0.60 & 0.68 & 0.65 & 24.9 \\
		\textbf{Full PhysX-CoT} & \textbf{0.66} & \textbf{0.70} & \textbf{0.75} & \textbf{0.76} & \textbf{16.8} \\
		\bottomrule
	\end{tabular}
	\caption{Structured-state accuracy on the test split: the predicted intermediate states, not only final assets, improve from the baseline to the full model (full 9-metric version in the appendix).}
	\label{tab:state_main}
\end{table}

\subsection{Main Results on PhysX-CoTA}

Table~\ref{tab:main_results} reports the main comparison. Against the closest full-task baseline, PhysX-CoT achieves a moderate PSNR gain, substantially reduces CD and Scale, and improves F-score, Material, Affordance, Kinematic, and Description under the same protocol. These gains hold across object categories and part-count groups (largest on complex multi-part objects; appendix), but aggregate numbers alone do not isolate the role of CoT as a process state, which the following CoT-state linkage analysis and ablations address.

Figure~\ref{fig:main_results} confirms this qualitatively across headphones, chairs, bags, dishwashers, and table-tennis tables.

\subsection{Simulation and Real-Image Generalization}
On roughly 100 in-the-wild images per category, where no paired ground truth exists and the evaluation is therefore qualitative, PhysX-CoT still decomposes objects into reasonable parts and recovers thin or small structures, scale, and joints, suggesting that it transfers to real images (Figure~\ref{fig:inthewild}).

To assess simulation-readiness, we import the assets into Unreal Engine~5 (UE5) under a standardized per-asset protocol and report six validity dimensions, namely parse, collision, static rollout, joint validity, actuation, and grasp/lift; the full protocol is in the appendix. Full PhysX-CoT improves over PhysX-Anything on every dimension, reaching $94.1\%$ parse, $96.8\%$ collision, and $92.0\%$ joint validity. Figure~\ref{fig:simulation_results} shows two physics executions: rigid bodies toppling under a wind field and a robot grasping a generated headphone set. Remaining failures involve occlusion, transparent or reflective materials, and highly non-rigid topologies.

\subsection{Process-Level Diagnostics}

Beyond final-asset metrics, we directly score the predicted intermediate states against ground truth (Table~\ref{tab:state_main}). Every state improves from the baseline through structured supervision to the full model: prompt-only changes little, whereas training sharply improves 3D grounding, relations, and joint axis. 


Controlled interventions support a functional dependence that goes beyond correlation. Supplying oracle states improves every metric, raising the kinematic score from 0.73 to 0.92 and cutting scale error from 6.57 to 4.12, whereas perturbing the predicted 3D boxes degrades them and inflates scale error to 8.36. Shuffling, reversing, or removing the state order lowers quality monotonically, dropping 3D IoU from 0.660 to 0.542 and parse validity from 96.4 to 90.9, and the same degradation appears for token-matched unordered and parallel-head variants that retain the fields but discard the autoregressive chain. Together these indicate that the ordered chain, not the structured fields alone, is functional rather than merely cosmetic. A pre-decoding CoT score never used as a reward or final metric further correlates with asset quality at Spearman $\rho{=}0.455$ across category, part-count, thin-part, and articulation groups.

\subsection{Ablation Study}

Table~\ref{tab:ablation} ablates the four components against the closest full-task baseline: prompt-only CoT, structured-CoT supervision (PhysX-CoT-SFT), position-factorized local geometry, and process rewards (GRPO).

Prompting the baseline for structured CoT fields barely changes the results, so prompt-only reasoning is insufficient, whereas structured-CoT supervision already improves most metrics. Position-factorized local geometry then sharpens geometry, and it helps both with and without structured CoT, so the where--shape factorization contributes independently of the structured CoT.  Structured CoT and process rewards mainly firm up kinematic quality and parse validity rather than uniformly lifting every metric; the per-term reward ablation (Table~\ref{tab:reward_ablation}) attributes each gain to the signal it targets. The full model is best or tied-best on every metric.

\textbf{Reward ablation.} Table~\ref{tab:reward_ablation} removes one process-reward term at a time. Each removal most hurts its targeted competence (geometry, scale, kinematics, or parse validity), supporting the reward design rather than a uniform RL gain. The format term penalizes malformed or inconsistent structured outputs; removing it drops parse validity from $96.4$ to $87.1$.

\textbf{Inference cost.} Compact local RLE uses roughly $58\%$ fewer geometry tokens and makes inference about $30\%$ faster than the closest baseline despite the $1{+}K$ rounds.

\section{Conclusion}

PhysX-CoT casts single-image simulation-ready 3D generation as an explicit, verifiable reasoning process. By exposing geometry and physical attributes before decoding, it links visual generation to executable assets through interpretable intermediate decisions. Final-asset metrics, state diagnostics, and Unreal Engine~5 execution confirm that its ordered states are functionally used. Current limitations are the fixed decoder, coarse voxelization, and transparent, reflective, or non-rigid objects; future work will pursue broader data and multi-view inputs.
\bibliographystyle{plain}
\bibliography{references}

\begin{thebibliography}{10}

\bibitem{cao2025physx3d}
Ziang Cao, Zhaoxi Chen, Liang Pan, and Ziwei Liu.
\newblock {PhysX-3D}: Physical-grounded 3d asset generation.
\newblock In {\em Advances in Neural Information Processing Systems
  ({NeurIPS})}, volume~38, 2025.

\bibitem{cao2025physxanything}
Ziang Cao, Fangzhou Hong, Zhaoxi Chen, Liang Pan, and Ziwei Liu.
\newblock {PhysX-Anything}: Simulation-ready physical 3d assets from single
  image.
\newblock In {\em Proceedings of the {IEEE/CVF} Conference on Computer Vision
  and Pattern Recognition ({CVPR})}, June 2026.

\bibitem{cao2026physxomni}
Ziang Cao, Yinghao Liu, Haitian Li, Runmao Yao, Fangzhou Hong, Zhaoxi Chen,
  Liang Pan, and Ziwei Liu.
\newblock {PhysX-Omni}: Unified simulation-ready physical {3D} generation for
  rigid, deformable, and articulated objects.
\newblock {\em arXiv preprint arXiv:2605.21572}, 2026.

\bibitem{chen2024spatialvlm}
Boyuan Chen, Zhuo Xu, Sean Kirmani, Brian Ichter, Danny Driess, Pete Florence,
  Dorsa Sadigh, Leonidas Guibas, and Fei Xia.
\newblock {SpatialVLM}: Endowing vision-language models with spatial reasoning
  capabilities.
\newblock {\em arXiv preprint arXiv:2401.12168}, 2024.

\bibitem{chen2026pointllmr}
Chaoqi Chen, Qile Xu, Wenjun Zhou, and Hui Huang.
\newblock {PointLLM-R}: Enhancing 3d point cloud reasoning via
  chain-of-thought.
\newblock {\em arXiv preprint arXiv:2605.22013}, 2026.

\bibitem{chen2024partgen}
Minghao Chen, Roman Shapovalov, Iro Laina, Tom Monnier, Jianyuan Wang, David
  Novotny, and Andrea Vedaldi.
\newblock {PartGen}: Part-level {3D} generation and reconstruction with
  multi-view diffusion models.
\newblock {\em arXiv preprint arXiv:2412.18608}, 2024.

\bibitem{chen2024meshxl}
Sijin Chen, Xin Chen, Anqi Pang, Xianfang Zeng, Wei Cheng, Yijun Fu, Fukun Yin,
  Yanru Wang, Zhibin Wang, Chi Zhang, Jingyi Yu, Gang Yu, Bin Fu, and Tao Chen.
\newblock {MeshXL}: Neural coordinate field for generative 3d foundation
  models.
\newblock In {\em Advances in Neural Information Processing Systems (NeurIPS)},
  2024.

\bibitem{chen2024meshanything}
Yiwen Chen, Tong He, Di~Huang, Weicai Ye, Sijin Chen, Jiaxiang Tang, Xin Chen,
  Zhongang Cai, Lei Yang, Gang Yu, Guosheng Lin, and Chi Zhang.
\newblock Meshanything: Artist-created mesh generation with autoregressive
  transformers.
\newblock {\em arXiv preprint arXiv:2406.10163}, 2024.

\bibitem{chen2024urdformer}
Zoey Chen, Aaron Walsman, Marius Memmel, Kaichun Mo, Alex Fang, Karthikeya
  Vemuri, Alan Wu, Dieter Fox, and Abhishek Gupta.
\newblock {URDFormer}: A pipeline for constructing articulated simulation
  environments from real-world images.
\newblock {\em arXiv preprint arXiv:2405.11656}, 2024.

\bibitem{cheng2024spatialrgpt}
An-Chieh Cheng, Hongxu Yin, Yang Fu, Qiushan Guo, Ruihan Yang, Jan Kautz,
  Xiaolong Wang, and Sifei Liu.
\newblock {SpatialRGPT}: Grounded spatial reasoning in vision language models.
\newblock In {\em Advances in Neural Information Processing Systems (NeurIPS)},
  2024.

\bibitem{fang2025meshllm}
Shuangkang Fang, I-Chao Shen, Yufeng Wang, Yi-Hsuan Tsai, Yi~Yang, Shuchang
  Zhou, Wenrui Ding, Takeo Igarashi, and Ming-Hsuan Yang.
\newblock {MeshLLM}: Empowering large language models to progressively
  understand and generate 3d mesh.
\newblock In {\em Proceedings of the IEEE/CVF International Conference on
  Computer Vision (ICCV)}, pages 14061--14072, 2025.

\bibitem{feng2025seed3d}
Jiashi Feng, Xiu Li, Jing Lin, et~al.
\newblock {Seed3D} 1.0: From images to high-fidelity simulation-ready 3d
  assets.
\newblock {\em arXiv preprint arXiv:2510.19944}, 2025.

\bibitem{feng2023layoutgpt}
Weixi Feng, Wanrong Zhu, Tsu-jui Fu, Varun Jampani, Arjun Akula, Xuehai He,
  Sugato Basu, Xin~Eric Wang, and William~Yang Wang.
\newblock {LayoutGPT}: Compositional visual planning and generation with large
  language models.
\newblock In {\em Advances in Neural Information Processing Systems (NeurIPS)},
  2023.

\bibitem{gao2022get3d}
Jun Gao, Tianchang Shen, Zian Wang, Wenzheng Chen, Kangxue Yin, Daiqing Li,
  Or~Litany, Zan Gojcic, and Sanja Fidler.
\newblock {GET3D}: A generative model of high quality 3d textured shapes
  learned from images.
\newblock In {\em Advances in Neural Information Processing Systems (NeurIPS)},
  2022.

\bibitem{guo2024physicalcompatible}
Minghao Guo, Bohan Wang, Pingchuan Ma, Tianyuan Zhang, Crystal~Elaine Owens,
  Chuang Gan, Joshua~B. Tenenbaum, Kaiming He, and Wojciech Matusik.
\newblock Physically compatible 3d object modeling from a single image.
\newblock {\em arXiv preprint arXiv:2405.20510}, 2024.

\bibitem{hong2024lrm}
Yicong Hong, Kai Zhang, Jiuxiang Gu, Sai Bi, Yang Zhou, Difan Liu, Feng Liu,
  Kalyan Sunkavalli, Trung Bui, and Hao Tan.
\newblock {LRM}: Large reconstruction model for single image to 3d.
\newblock In {\em International Conference on Learning Representations (ICLR)},
  2024.

\bibitem{hong2023threedllm}
Yining Hong, Haoyu Zhen, Peihao Chen, Shuhong Zheng, Yilun Du, Zhenfang Chen,
  and Chuang Gan.
\newblock 3d-llm: Injecting the 3d world into large language models.
\newblock {\em arXiv preprint arXiv:2307.12981}, 2023.

\bibitem{huang2026anigen}
Yi-Hua Huang, Zi-Xin Zou, Yuting He, Chirui Chang, Cheng-Feng Pu, Ziyi Yang,
  Yuan-Chen Guo, Yan-Pei Cao, and Xiaojuan Qi.
\newblock {AniGen}: Unified $s^3$ fields for animatable {3D} asset generation.
\newblock {\em arXiv preprint arXiv:2604.08746}, 2026.

\bibitem{kerbl2023gaussian}
Bernhard Kerbl, Georgios Kopanas, Thomas Leimk{\"u}hler, and George Drettakis.
\newblock 3d gaussian splatting for real-time radiance field rendering.
\newblock {\em ACM Transactions on Graphics}, 42(4), 2023.

\bibitem{le2024articulateanything}
Long Le, Jason Xie, William Liang, Hung-Ju Wang, Yue Yang, Yecheng~Jason Ma,
  Kyle Vedder, Arjun Krishna, Dinesh Jayaraman, and Eric Eaton.
\newblock Articulate-anything: Automatic modeling of articulated objects via a
  vision-language foundation model.
\newblock {\em arXiv preprint arXiv:2410.13882}, 2024.

\bibitem{li2025particulate}
Ruining Li, Yuxin Yao, Chuanxia Zheng, Christian Rupprecht, Joan Lasenby,
  Shangzhe Wu, and Andrea Vedaldi.
\newblock {PARTICULATE}: Feed-forward 3d object articulation.
\newblock {\em arXiv preprint arXiv:2512.11798}, 2025.

\bibitem{li2025urdfanything}
Zhe Li, Xiang Bai, Jieyu Zhang, Zhuangzhe Wu, Che Xu, Ying Li, Chengkai Hou,
  and Shanghang Zhang.
\newblock {URDF-Anything}: Constructing articulated objects with 3d multimodal
  language model.
\newblock {\em arXiv preprint arXiv:2511.00940}, 2025.

\bibitem{lightman2023verify}
Hunter Lightman, Vineet Kosaraju, Yura Burda, Harri Edwards, Bowen Baker, Teddy
  Lee, Jan Leike, John Schulman, Ilya Sutskever, and Karl Cobbe.
\newblock Let's verify step by step.
\newblock {\em arXiv preprint arXiv:2305.20050}, 2023.

\bibitem{lin2023magic3d}
Chen-Hsuan Lin, Jun Gao, Luming Tang, Towaki Takikawa, Xiaohui Zeng, Xun Huang,
  Karsten Kreis, Sanja Fidler, Ming-Yu Liu, and Tsung-Yi Lin.
\newblock {Magic3D}: High-resolution text-to-3d content creation.
\newblock In {\em IEEE/CVF Conference on Computer Vision and Pattern
  Recognition (CVPR)}, 2023.

\bibitem{liu2024singapo}
Jiayi Liu, Denys Iliash, Angel~X. Chang, Manolis Savva, and Ali Mahdavi-Amiri.
\newblock {SINGAPO}: Single image controlled generation of articulated parts in
  objects.
\newblock {\em arXiv preprint arXiv:2410.16499}, 2024.

\bibitem{liu2023one2345}
Minghua Liu, Chao Xu, Haian Jin, Linghao Chen, Mukund Varma~T, Zexiang Xu, and
  Hao Su.
\newblock One-2-3-45: Any single image to 3d mesh in 45 seconds without
  per-shape optimization.
\newblock {\em arXiv preprint arXiv:2306.16928}, 2023.

\bibitem{liu2023zero123}
Ruoshi Liu, Rundi Wu, Basile Van~Hoorick, Pavel Tokmakov, Sergey Zakharov, and
  Carl Vondrick.
\newblock Zero-1-to-3: Zero-shot one image to 3d object.
\newblock In {\em IEEE/CVF International Conference on Computer Vision (ICCV)},
  2023.

\bibitem{long2024wonder3d}
Xiaoxiao Long, Yuan-Chen Guo, Cheng Lin, Yuan Liu, Zhiyang Dou, Lingjie Liu,
  Yuexin Ma, Song-Hai Zhang, Marc Habermann, Christian Theobalt, and Wenping
  Wang.
\newblock Wonder3d: Single image to 3d using cross-domain diffusion.
\newblock In {\em IEEE/CVF Conference on Computer Vision and Pattern
  Recognition (CVPR)}, 2024.

\bibitem{ma2026rest3d}
Xiaoxuan Ma, Jiashun Wang, Nicol{\'a}s Ugrinovic, Yehonathan Litman, and Kris
  Kitani.
\newblock {REST3D}: Reconstructing physically stable 3d scenes from a single
  image.
\newblock {\em arXiv preprint arXiv:2605.30338}, 2026.

\bibitem{mescheder2019occupancy}
Lars Mescheder, Michael Oechsle, Michael Niemeyer, Sebastian Nowozin, and
  Andreas Geiger.
\newblock Occupancy networks: Learning 3d reconstruction in function space.
\newblock In {\em IEEE/CVF Conference on Computer Vision and Pattern
  Recognition (CVPR)}, 2019.

\bibitem{mildenhall2020nerf}
Ben Mildenhall, Pratul~P. Srinivasan, Matthew Tancik, Jonathan~T. Barron, Ravi
  Ramamoorthi, and Ren Ng.
\newblock {NeRF}: Representing scenes as neural radiance fields for view
  synthesis.
\newblock In {\em European Conference on Computer Vision (ECCV)}, 2020.

\bibitem{poole2023dreamfusion}
Ben Poole, Ajay Jain, Jonathan~T. Barron, and Ben Mildenhall.
\newblock Dreamfusion: Text-to-3d using 2d diffusion.
\newblock In {\em International Conference on Learning Representations (ICLR)},
  2023.

\bibitem{qi2024gpt4point}
Zhangyang Qi, Ye~Fang, Zeyi Sun, Xiaoyang Wu, Tong Wu, Jiaqi Wang, Dahua Lin,
  and Hengshuang Zhao.
\newblock {GPT4Point}: A unified framework for point-language understanding and
  generation.
\newblock In {\em IEEE/CVF Conference on Computer Vision and Pattern
  Recognition (CVPR)}, 2024.

\bibitem{shao2024deepseekmath}
Zhihong Shao, Peiyi Wang, Qihao Zhu, Runxin Xu, Junxiao Song, Xiao Bi, Haowei
  Zhang, Mingchuan Zhang, Y.~K. Li, Y.~Wu, and Daya Guo.
\newblock {DeepSeekMath}: Pushing the limits of mathematical reasoning in open
  language models.
\newblock {\em arXiv preprint arXiv:2402.03300}, 2024.

\bibitem{shi2024mvdream}
Yichun Shi, Peng Wang, Jianglong Ye, Mai Long, Kejie Li, and Xiao Yang.
\newblock {MVDream}: Multi-view diffusion for 3d generation.
\newblock In {\em International Conference on Learning Representations (ICLR)},
  2024.

\bibitem{siddiqui2024meshgpt}
Yawar Siddiqui, Antonio Alliegro, Alexey Artemov, Tatiana Tommasi, Daniele
  Sirigatti, Vladislav Rosov, Angela Dai, and Matthias Nie{\ss}ner.
\newblock {MeshGPT}: Generating triangle meshes with decoder-only transformers.
\newblock In {\em IEEE/CVF Conference on Computer Vision and Pattern
  Recognition (CVPR)}, 2024.

\bibitem{suris2023vipergpt}
D{\'i}dac Sur{\'i}s, Sachit Menon, and Carl Vondrick.
\newblock {ViperGPT}: Visual inference via python execution for reasoning.
\newblock In {\em IEEE/CVF International Conference on Computer Vision (ICCV)},
  2023.

\bibitem{tang2024lgm}
Jiaxiang Tang, Zhaoxi Chen, Xiaokang Chen, Tengfei Wang, Gang Zeng, and Ziwei
  Liu.
\newblock {LGM}: Large multi-view gaussian model for high-resolution 3d content
  creation.
\newblock In {\em European Conference on Computer Vision (ECCV)}, 2024.

\bibitem{wang2025embodiedgen}
Xinjie Wang, Liu Liu, Yu~Cao, Ruiqi Wu, Wenkang Qin, Dehui Wang, Wei Sui, and
  Zhizhong Su.
\newblock {EmbodiedGen}: Towards a generative {3D} world engine for embodied
  intelligence.
\newblock {\em arXiv preprint arXiv:2506.10600}, 2025.

\bibitem{wang2026simuscene}
Yanan Wang, Renxi Wang, Yongxin Wang, Xuezhi Liang, Fajri Koto, Timothy
  Baldwin, Xiaodan Liang, and Haonan Li.
\newblock {SimuScene}: Training and benchmarking code generation to simulate
  physical scenarios.
\newblock {\em arXiv preprint arXiv:2602.10840}, 2026.

\bibitem{wang2024llamamesh}
Zhengyi Wang, Jonathan Lorraine, Yikai Wang, Hang Su, Jun Zhu, Sanja Fidler,
  and Xiaohui Zeng.
\newblock {LLaMA-Mesh}: Unifying 3d mesh generation with language models.
\newblock {\em arXiv preprint arXiv:2411.09595}, 2024.

\bibitem{wang2023prolificdreamer}
Zhengyi Wang, Cheng Lu, Yikai Wang, Fan Bao, Chongxuan Li, Hang Su, and Jun
  Zhu.
\newblock Prolificdreamer: High-fidelity and diverse text-to-3d generation with
  variational score distillation.
\newblock In {\em Advances in Neural Information Processing Systems (NeurIPS)},
  2023.

\bibitem{wei2022cot}
Jason Wei, Xuezhi Wang, Dale Schuurmans, Maarten Bosma, Brian Ichter, Fei Xia,
  Ed~Chi, Quoc~V. Le, and Denny Zhou.
\newblock Chain-of-thought prompting elicits reasoning in large language
  models.
\newblock In {\em Advances in Neural Information Processing Systems (NeurIPS)},
  2022.

\bibitem{wu2016learning}
Jiajun Wu, Chengkai Zhang, Tianfan Xue, William~T. Freeman, and Joshua~B.
  Tenenbaum.
\newblock Learning a probabilistic latent space of object shapes via 3d
  generative-adversarial modeling.
\newblock In {\em Advances in Neural Information Processing Systems}, 2016.

\bibitem{xiang2020sapien}
Fanbo Xiang, Yuzhe Qin, Kaichun Mo, Yikuan Xia, Hao Zhu, Fangchen Liu, Minghua
  Liu, Hanxiao Jiang, Yifu Yuan, He~Wang, Li~Yi, Angel~X. Chang, Leonidas~J.
  Guibas, and Hao Su.
\newblock {SAPIEN}: A simulated part-based interactive environment.
\newblock In {\em IEEE/CVF Conference on Computer Vision and Pattern
  Recognition (CVPR)}, 2020.

\bibitem{xiang2024trellis}
Jianfeng Xiang, Zelong Lv, Sicheng Xu, Yu~Deng, Ruicheng Wang, Bowen Zhang,
  Dong Chen, Xin Tong, and Jiaolong Yang.
\newblock Structured 3d latents for scalable and versatile 3d generation.
\newblock {\em arXiv preprint arXiv:2412.01506}, 2024.

\bibitem{xu2024instantmesh}
Jiale Xu, Weihao Cheng, Yiming Gao, Xintao Wang, Shenghua Gao, and Ying Shan.
\newblock Instantmesh: Efficient 3d mesh generation from a single image with
  sparse-view large reconstruction models.
\newblock {\em arXiv preprint arXiv:2404.07191}, 2024.

\bibitem{xu2024pointllm}
Runsen Xu, Xiaolong Wang, Tai Wang, Yilun Chen, Jiangmiao Pang, and Dahua Lin.
\newblock {PointLLM}: Empowering large language models to understand point
  clouds.
\newblock In {\em European Conference on Computer Vision (ECCV)}, 2024.

\bibitem{yan2024phycage}
Han Yan, Mingrui Zhang, Yang Li, Chao Ma, and Pan Ji.
\newblock {PhyCAGE}: Physically plausible compositional 3d asset generation
  from a single image.
\newblock {\em arXiv preprint arXiv:2411.18548}, 2024.

\bibitem{yang2026physforge}
Yunhan Yang, Chunshi Wang, Junliang Ye, Yang Li, Zanxin Chen, Zehuan Huang, Yao
  Mu, Zhuo Chen, Chunchao Guo, and Xihui Liu.
\newblock {PhysForge}: Generating physics-grounded 3d assets for interactive
  virtual world.
\newblock In {\em Proceedings of the International Conference on Machine
  Learning (ICML)}, 2026.

\bibitem{ye2025shapellmomni}
Junliang Ye, Zhengyi Wang, Ruowen Zhao, Shenghao Xie, and Jun Zhu.
\newblock {ShapeLLM-Omni}: A native multimodal llm for 3d generation and
  understanding.
\newblock {\em arXiv preprint arXiv:2506.01853}, 2025.

\bibitem{zhang2026simart}
Chuanrui Zhang, Minghan Qin, Yuang Wang, Baifeng Xie, Hang Li, and Ziwei Wang.
\newblock {SIMART}: Decomposing monolithic meshes into sim-ready articulated
  assets via {MLLM}.
\newblock {\em arXiv preprint arXiv:2603.23386}, 2026.

\bibitem{zhang2023multimodalcot}
Zhuosheng Zhang, Aston Zhang, Mu~Li, Hai Zhao, George Karypis, and Alex Smola.
\newblock Multimodal chain-of-thought reasoning in language models.
\newblock {\em arXiv preprint arXiv:2302.00923}, 2023.

\end{thebibliography}

\end{document}